\documentclass[sigconf,nonacm]{acmart}

\AtBeginDocument{%
  }

\setcopyright{rightsretained}
\copyrightyear{2026}
\acmYear{2026}
\acmDOI{}
\acmISBN{}

\usepackage{booktabs}
\usepackage{graphicx}
\usepackage{amsmath}
\usepackage{tikz}
\usetikzlibrary{arrows.meta,positioning}
\usepackage{listings}
\usepackage{newunicodechar}
\newunicodechar{→}{\ensuremath{\rightarrow}}
\newunicodechar{←}{\ensuremath{\leftarrow}}
\newunicodechar{↔}{\ensuremath{\leftrightarrow}}
\newunicodechar{↓}{\ensuremath{\downarrow}}
\newunicodechar{↑}{\ensuremath{\uparrow}}
\newunicodechar{≈}{\ensuremath{\approx}}
\newunicodechar{×}{\ensuremath{\times}}
\newunicodechar{÷}{\ensuremath{\div}}
\newunicodechar{≥}{\ensuremath{\geq}}
\newunicodechar{≤}{\ensuremath{\leq}}
\newunicodechar{±}{\ensuremath{\pm}}
\newunicodechar{−}{\ensuremath{-}}
\newunicodechar{•}{\textbullet}
\newunicodechar{°}{\ensuremath{^\circ}}
\newunicodechar{…}{\ldots}

\begin{document}

\title{Retrieve, Match, Escalate: Accurate and Scalable Product Linking with VLM-Distilled Cross-Encoders and Agentic VLMs}

\author{Jian Wang}
\email{jian.wang@doordash.com}
\affiliation{%
  \institution{DoorDash Inc.}
  \city{San Francisco}
  \state{CA}
  \country{USA}}
\author{Steven Xu}
\email{steven.xu@doordash.com}
\affiliation{%
  \institution{DoorDash Inc.}
  \city{San Francisco}
  \state{CA}
  \country{USA}}
\author{Sanjyot Thete}
\email{sanjyot.thete@doordash.com}
\affiliation{%
  \institution{DoorDash Inc.}
  \city{San Francisco}
  \state{CA}
  \country{USA}}
\author{Maryam Barouti}
\email{maryam.barouti@doordash.com}
\affiliation{%
  \institution{DoorDash Inc.}
  \city{San Francisco}
  \state{CA}
  \country{USA}}
\author{Tom Tang}
\email{tom.tang@doordash.com}
\affiliation{%
  \institution{DoorDash Inc.}
  \city{San Francisco}
  \state{CA}
  \country{USA}}
\author{Elaine Wu}
\email{elaine.wu@doordash.com}
\affiliation{%
  \institution{DoorDash Inc.}
  \city{San Francisco}
  \state{CA}
  \country{USA}}
\author{Charu Sareen}
\email{charu.sareen@doordash.com}
\affiliation{%
  \institution{DoorDash Inc.}
  \city{San Francisco}
  \state{CA}
  \country{USA}}
\author{Kyle MacDonald}
\email{kyle.macdonald@doordash.com}
\affiliation{%
  \institution{DoorDash Inc.}
  \city{San Francisco}
  \state{CA}
  \country{USA}}
\renewcommand{\shortauthors}{Wang et al.}

\begin{abstract}
Product linking, the entity-resolution task of mapping merchant product records to canonical catalog products, consolidates fragmented listings so downstream search, recommendation, and advertising see one clean entry per product. At marketplace scale, billions of noisy, multi-category records must be resolved against tens of millions of canonical products, where scoring every candidate with a single model is either too weak for the hard cases or too costly for the easy ones. We present a production retrieve-then-match cascade that spends computation in proportion to difficulty: retrieval surfaces plausible matches, a lightweight text cross-encoder auto-resolves the high-confidence majority, and an agentic multimodal vision-language model settles the ambiguous remainder by inspecting product images and issuing web searches for evidence that is in neither record. The cross-encoder is distilled from millions of dual-VLM-consensus labels, retiring human annotation from the training set, and is calibrated to auto-accept links at a 98\% precision bar validated against a smaller operator-certified audit. The agent is a self-hosted open-weight model that reaches a closed frontier VLM's precision at a four-point recall cost (88\% versus 92\%) for roughly one-seventh the per-pair cost, with no fine-tuning. Per-pair cost spans nearly five orders of magnitude from the cheap cross-encoder to the frontier VLM, so escalating only the hard tail to the agent raises end-to-end link coverage from the cheap stage's 68\% to 77\%.
\end{abstract}


\maketitle

\hypersetup{%
  pdfauthor={Jian Wang, Steven Xu, Sanjyot Thete, Maryam Barouti, Tom Tang, Elaine Wu, Charu Sareen, Kyle MacDonald},
  pdftitle={Retrieve, Match, Escalate: Accurate and Scalable Product Linking with VLM-Distilled Cross-Encoders and Agentic VLMs}}
\pdfinfo{/Author (Jian Wang, Steven Xu, Sanjyot Thete, Maryam Barouti, Tom Tang, Elaine Wu, Charu Sareen, Kyle MacDonald)}

\section{Introduction}

Product linking maps merchant product records, or merchant SKUs, to
the corresponding canonical products in a marketplace catalog.
Because independent merchants do not always represent the same
product using the same names, identifiers, attributes, or images,
equivalent listings may appear as distinct records. Linking these
records to a shared canonical entity allows product metadata and
behavioral signals to be pooled across listings, yielding more
complete item representations for downstream search,
recommendation, and advertising.

At marketplace scale, this requires resolving billions
of noisy, multilingual, and multi-category merchant records against
tens of millions of canonical products while preserving
variant-level distinctions such as size and flavor.
False links merge distinct products and contaminate the signals
associated with them, whereas missed links leave equivalent listings
fragmented and prevent information from being shared.

Product linking naturally decomposes into retrieval and matching.
For each merchant SKU, the system first retrieves a small set of
plausible canonical products and then determines whether any
candidate represents the same underlying product. Candidate pairs
vary substantially in difficulty. Strong agreement in identifiers,
names, or other catalog fields can make some pairs straightforward,
whereas missing, noisy, or conflicting evidence makes others
ambiguous. Applying the same model uniformly therefore creates a
cost-capability trade-off: it is either wasteful on easy pairs or
inadequate for the hard tail.

Recent work addresses this trade-off by allocating computation according to difficulty, resolving straightforward cases with lightweight models and escalating ambiguous cases to more capable reasoning systems~\cite{li2025arter,chen2023frugalgpt}. The closest work applies adaptive routing to text-based entity linking, where reasoning is performed over evidence already provided to the model. Product linking presents additional challenges. Ambiguous pairs may require reasoning over product images or retrieving evidence absent
from both catalog records. Reliable training labels are also
difficult to obtain because product identity is conjunctive: a valid
match must be consistent across several attributes that distinguish
product variants (e.g., brand, product line, size), while a mismatch in any one can indicate a different canonical product. Annotators must therefore jointly resolve these attributes from incomplete or conflicting evidence, making human labeling difficult
to scale.

In this paper, we propose an end-to-end product-linking pipeline
that combines the efficiency of a cross-encoder with the multimodal
reasoning capabilities of a VLM agent through a confidence-routed
model cascade. After the retrieval stage, a cross-encoder trained on
dual-VLM-consensus labels scores every candidate pair.
High-confidence matches and non-matches are accepted or rejected
directly, while uncertain pairs are deferred to a tool-using VLM
agent that performs deeper reasoning across modalities and, when
needed, searches the web for additional evidence.

Our core contributions are threefold. First, we formulate ambiguous
product linking as a tool-using multimodal inference task and
implement it using a pretrained VLM with task-specific prompting,
multimodal product context, and access to web search. Second, we
demonstrate how to make agentic multimodal product linking practical
at marketplace scale by pairing the VLM agent with a lightweight
cross-encoder in a confidence-routed cascade. Trained on
dual-VLM-consensus labels, the cross-encoder resolves routine cases
efficiently while reserving agentic reasoning for uncertain pairs.
Third, we validate the pipeline at a production operating point and
report the findings behind it: the open-weight, self-hosted agent
reaches precision parity with a closed frontier model at a fraction
of the cost with no fine-tuning, and AI-agent-driven experimentation
(autoresearch) surfaces the accuracy drivers, including a
representation that wins in aggregate yet fails on a rare adversarial
sub-distribution and is addressed by a second distillation pass.

\section{Related Work}

\paragraph{Product matching and entity resolution.}
Deciding whether two records denote the same real-world entity is the classical problem of entity resolution, which formalizes the blocking-then-matching pipeline our retrieve-then-match funnel inherits~\cite{christen2012}. Learned matchers have progressively replaced hand-tuned rules: Magellan~\cite{konda2016magellan} provides an end-to-end matching toolkit; Ditto~\cite{li2020ditto} casts pairwise matching as fine-tuned Transformer sequence-pair classification, the direct ancestor of our text-only cross-encoder; and the WDC benchmark~\cite{peeters2024wdc} establishes product matching over noisy web data as a canonical evaluation. Closest in framing are the multimodal fashion matcher of T\'{o}th et al.~\cite{toth2024multimodal}, which fuses frozen image and text encoders into one contrastive space and reaches production precision through human-in-the-loop validation, and an industrial deduplication pipeline that resolves candidates with a single decider model~\cite{hepsiburada2025dedup}. Both resolve every candidate pair with a single, non-adaptive decision. We instead route by confidence: a text-only cross-encoder resolves the high-signal majority, and only an ambiguous minority escalates to an agent that actively gathers evidence. Where text-only matching trails image embeddings in fashion~\cite{toth2024multimodal}, our broad-catalog matcher resolves most pairs from raw barcode tokens alone.

\paragraph{Adaptive routing and cost-tiered inference.}
Spending computation only on hard inputs is long-standing, from the attentional cascade of Viola and Jones~\cite{violajones2001} and Chow's reject option~\cite{chow1970reject} to FrugalGPT~\cite{chen2023frugalgpt}, which cascades from cheap to expensive language models at fixed quality. The closest prior art is ARTER~\cite{li2025arter}, which applies confidence-based adaptive routing to entity linking: a cheap frozen linker resolves easy mentions, and only hard mentions are escalated to targeted large-language-model reasoning. We share its retrieve-then-match structure and its thesis that not all instances need expensive reasoning, but differ on three axes. Our escalation tier is multimodal and agentic, seeing product images and issuing multi-round tool calls rather than a single-shot text-only prompt over a fixed candidate list; our routing uses the cheap matcher's own calibrated score under band thresholds, with no separately trained router; and the cheap tier is itself distilled from the expensive tier rather than assembled from frozen models. We also report a production operating point, recall at high precision over billions of records, rather than an offline benchmark. Confidence-based escalation is being applied concurrently to related e-commerce and agentic tasks, from attribute-quality assessment in product catalogs~\cite{amazon2025cascade} to routing computer-use-agent actions to the cheapest adequate model~\cite{avr2026routing}.

\section{System Architecture}
\label{sec:overview}

\subsection{Overview}

The platform ingests \emph{merchant product records}: listings supplied by a large, heterogeneous population of independent merchants, each carrying a name, one or more images, structured attributes such as size, unit, and barcodes, and a coarse product type. It also maintains a \emph{canonical catalog product}: a single, deduplicated, quality-controlled representation of a distinct consumer product, numbering in the tens of millions. Linking answers, for each merchant record, which canonical product (if any) denotes the same consumer product, so that many noisy listings of one item share a clean, enriched entry. Because canonical products are themselves deduplicated over time, a record's correct target may be a successor of a since-merged product. Following the entity-resolution literature~\cite{christen2012,konda2016magellan,li2020ditto,li2025arter}, we frame linking as retrieve-then-match: given a record $r$, retrieval returns a small candidate set $C(r)$, and a match stage evaluates each $c \in C(r)$ with a pairwise binary decision $f(r,c)\in\{\text{same},\text{different}\}$, linking the record to a candidate judged the same. Recall is upper-bounded by retrieval; precision and cost are dominated by how the pairwise decisions are made and how aggressively they are automated. The same core serves an online build pipeline (linking records as they arrive) and offline post-build deduplication (re-examining existing links in bulk), over noisy, multilingual, multi-category merchant data (Section~\ref{sec:discussion}).

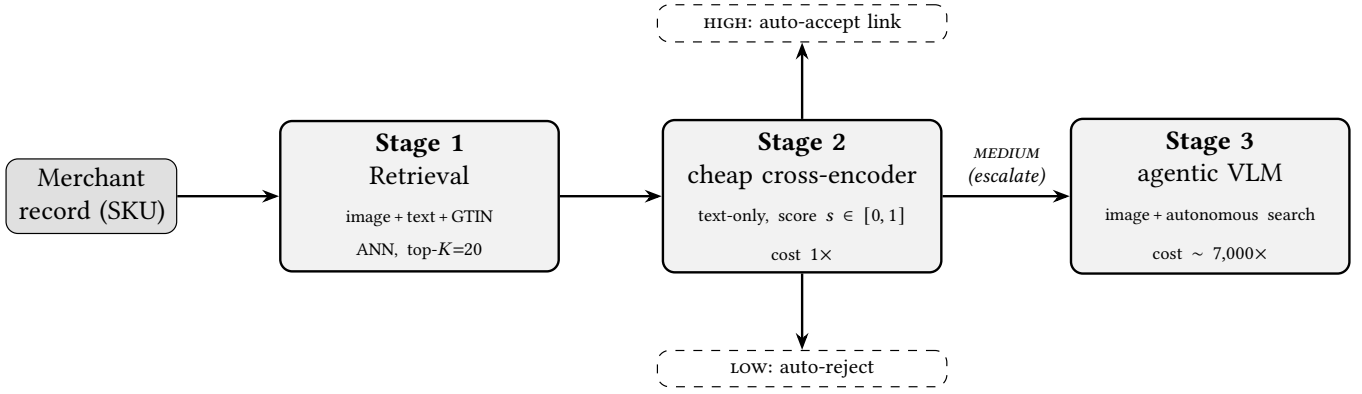
\begin{figure*}[t]
\centering
\resizebox{\textwidth}{!}{%
\begin{tikzpicture}[
  >={Stealth[length=2mm]},
  node distance=9mm and 12mm,
  stage/.style={rounded corners, draw, thick, fill=black!5, align=center, inner sep=4pt, minimum height=17mm, text width=30mm},
  io/.style={rounded corners, draw, fill=black!12, align=center, inner sep=3pt, text width=18mm},
  band/.style={rounded corners, draw, dashed, align=center, inner sep=3pt, text width=32mm, font=\footnotesize},
  lbl/.style={font=\footnotesize\itshape}
]
\node[io] (rec) {Merchant record (SKU)};
\node[stage, right=of rec] (s1) {\textbf{Stage 1}\\ Retrieval\\[2pt]{\scriptsize image\,+\,text\,+\,GTIN ANN, top-$K{=}20$}};
\node[stage, right=of s1] (s2) {\textbf{Stage 2}\\ cheap cross-encoder\\[2pt]{\scriptsize text-only, score $s\in[0,1]$}\\[2pt]{\scriptsize cost $1\times$}};
\node[stage, right=15mm of s2] (s3) {\textbf{Stage 3}\\ agentic VLM\\[2pt]{\scriptsize image\,+\,autonomous search}\\[2pt]{\scriptsize cost $\sim\!7{,}000\times$}};
\node[band, above=of s2] (hi) {\textsc{high}: auto-accept link};
\node[band, below=of s2] (lo) {\textsc{low}: auto-reject};
\draw[->,thick] (rec) -- (s1);
\draw[->,thick] (s1) -- (s2);
\draw[->,thick] (s2) -- (hi);
\draw[->,thick] (s2) -- (lo);
\draw[->,thick] (s2) -- node[lbl,above,align=center]{\textsc{medium}\\(escalate)} (s3);
\end{tikzpicture}%
}
\caption{The adaptive-routing cascade: retrieval narrows each merchant record to $K{=}20$ candidates, a cheap text-only cross-encoder routes each pair by confidence (auto-accepting \textsc{high}, auto-rejecting \textsc{low}), and only the ambiguous \textsc{medium} band escalates to the agentic multimodal VLM. Per-pair cost is relative to the cross-encoder (Table~\ref{tab:ladder}).}
\label{fig:funnel}
\end{figure*}

At our scale (billions of records, each expanded by retrieval into $K$ further candidate pairs), applying a heavyweight multimodal reasoner to every pair is infeasible, yet a cheap model alone leaves the ambiguous tail unresolved. We therefore route by difficulty, spending compute in proportion to how hard a decision is~\cite{chen2023frugalgpt,chow1970reject,li2025arter}. Figure~\ref{fig:funnel} shows the three stages: Stage~1 (retrieval) returns the top $K{=}20$ candidates per record from image, text, and barcode nearest-neighbor search over a vector index; Stage~2 is a cheap, text-only cross-encoder that scores each candidate pair and routes it by confidence; and Stage~3 is an agentic multimodal VLM that adjudicates only the pairs Stage~2 cannot resolve. We describe each stage below, then the routing policy that composes them (Section~\ref{sec:routing}).

\subsection{Retrieval}
\label{sec:retrieval}

Before any pair is scored, the system must narrow the global catalog to a small candidate set per record. This candidate-generation step is the classical blocking stage of entity resolution \cite{christen2012,konda2016magellan}, and its recall upper-bounds the recall of the entire funnel: a correct match that is never retrieved cannot be recovered by any downstream matcher.

Retrieval fuses three complementary channels over a vector index. A text approximate-nearest-neighbor (ANN) index embeds item names and structured fields; an image ANN index embeds product photographs \cite{radford2021clip}; and a barcode (GTIN) channel exploits exact and partial global-trade-identifier matches. The three channels are merged into a single top-$K$ candidate list, with $K = 20$ canonical products per record.

Several levers are under active evaluation to raise this ceiling. (i) Pushing attribute filtering (brand, coarse category, purchase-type) upstream to prune incompatible candidates before scoring. (ii) Increasing $K$ to trade recall against downstream cost. (iii) Stronger, multimodal embeddings, replacing the current general-purpose text embedding with jointly trained image-text encoders (Qwen3-VL and Gemini Embedding 2) \cite{toth2024multimodal}. (iv) Lexical retrieval (BM25/Jaccard) to complement dense ANN on rare tokens and long-tail names. (v) Dedicated UPC/PLU/GTIN retrieval to fully exploit barcode signals. (vi) Broader image coverage, since many records arrive without a usable photograph.

A single difficulty cuts across every lever: retrieval must ingest heterogeneous, multilingual, and frequently missing merchant-supplied fields (free-text sizes, un-normalized barcodes, placeholder or absent images, and names spanning many languages and scripts). We return to the implications of this input noise in the Discussion (Section~\ref{sec:discussion}).

\subsection{Cross-Encoder Matcher}
\label{sec:stage2}

The second stage decides, for each candidate pair surfaced by retrieval, whether a merchant product record and a canonical catalog product denote the same consumer product. Because most pairs carry strong, unambiguous signal (an exact barcode agreement, a near-identical name), the design goal for this stage is calibrated confidence rather than peak accuracy. The matcher's score can be thresholded into three bands (HIGH auto-accept, LOW auto-reject, MEDIUM escalate) so that only the ambiguous tail is forwarded to the expensive multimodal stage. We show that a compact, text-only cross-encoder, distilled from VLM judgments, is calibrated well enough to auto-resolve the high-signal majority and reduce the volume escalated to the expensive stage.

\subsubsection{Architecture}
The matcher is a ModernBERT-base cross-encoder~\cite{warner2024modernbert} with a sigmoid classification head, 150M parameters, and no image input. Both sides of a candidate pair are serialized and concatenated into a single sequence of at most 256 tokens, so that full cross-attention operates over the two products jointly rather than encoding each side into an independent embedding. This pairwise formulation lets the model reason about agreement between fields, for example, that two differently phrased names describe the same flavor and size. Each side contributes a small set of fields: a coarse category tag (\texttt{[FRESH]} or \texttt{[CPG]}), the item name, a GTIN list (up to five), a UPC list (up to five), a size string, a weighted-item flag, and a unit. We feed raw barcode digits directly into the sequence rather than precomputed equality or intersection flags. The model learns barcode agreement, and partial agreement, from the tokens themselves. We found this preferable on aggregate accuracy, though with a representation-specific caveat detailed in Section~\ref{sec:stage2gtin}. The result is a single small model that runs on commodity accelerators with no external calls.

\subsubsection{Labeling and Training Data}
We reuse the expensive Stage-3 VLMs as labelers, distilling their judgments into the cheap cross-encoder~\cite{hinton2015distilling}. We construct 5.3M training pairs primarily by what we call dual-VLM consensus: two different VLMs independently score each pair. We keep the 87\% on which they agree and drop the 13\% on which they disagree as label-ambiguous. Multi-model agreement has been used to filter pseudo-labels and to verify predictions at inference time~\cite{vlmcpl2024,ceocr2025consensus}; we instead treat two independent VLMs' agreement as the training label itself, unlike distilling a single teacher's rationales onto human-labeled ground truth~\cite{agrawal2025rationale}. Disagreement is a reliable proxy for intrinsic difficulty, and discarding it yields a training signal far more self-consistent than single-annotator labels, an effect consistent with LLM-as-judge calibration findings~\cite{buildjudgeoptimize2026}. We augment this with VLM labels on medium-confidence production predictions, hold out a zero-leak 6k-row development set, and mine negatives 50/50 hard/random against a 50/50 positive/negative balance so the operating threshold is not dominated by easy negatives. This replaces 40k human labels with 5.3M higher-consistency VLM labels, a $130\times$ scale-up. The failure mode is inherited bias: the student can only be as correct as the consensus of its teachers. Systematic VLM errors, and the ambiguous cases they silently drop, are invisible to the student and must be caught by the ops-certified audit (Section~\ref{sec:stage2results}) rather than by the distillation loss.

\subsection{VLM Agent}
\label{sec:stage3}

The cross-encoder matcher of Section~\ref{sec:stage2} auto-resolves most pairs, but it is deliberately text-only: when a merchant record carries no barcode, a placeholder or missing image, or a size buried in free text, the textual evidence alone cannot separate a true match from a near-duplicate. These are exactly the pairs the matcher routes to its MEDIUM band. Stage~3 is a multimodal agent built for this hard tail. Resolving these pairs often turns on evidence in neither record (a barcode or brand verifiable only against an external source), so the stage is agentic, gathering evidence over several steps rather than scoring in a single pass. It is the most capable and most expensive stage of the cascade, so it runs on the smallest slice of traffic; the design goal is to preserve, at reduced cost, the accuracy that only an agent with full information access can reach.

\subsubsection{The Task: A Tool-Using Multimodal Agent}
Stage~3 is invoked per candidate pair. The agent receives the two product records together with their images, and it may autonomously search the open web through a single tool exposed over a model-context-protocol (MCP) interface. It decides, on each reasoning step, whether it has enough evidence or whether to gather more, then emits a structured verdict as JSON, \{\texttt{reasoning}, \texttt{match\_decision}\}, with no scalar confidence score. Unlike Stage 2, which emits a calibrated $[0,1]$ score for three-band routing, Stage 3 has no confidence tier of its own. Throughout this section, ``medium confidence'' describes the upstream Stage-2 score of the input pair being escalated, not any property of the VLM's own binary output. This interleaving of free-form reasoning and tool calls follows the ReAct pattern \cite{yao2023react}, and the learned decision of when and how to invoke an external tool follows tool-augmented language modeling \cite{schick2023toolformer}. The design mirrors a human operator, who would inspect the packaging photo and search a retailer site for the barcode or brand; giving the model the image plus grounded search puts it on equal footing.

\begin{figure}[t]
\centering
\fbox{\begin{minipage}{0.90\columnwidth}
\footnotesize
\textbf{Linking-agent prompt (abridged).} You are given two product records. The item name is always present; image, GTIN/UPC barcode, size, unit, and a weighted-item flag may be missing. Decide whether the two records are the same consumer product.\\[2pt]
\textit{Selected rules.} A missing attribute is ``unknown,'' never evidence against a match. A generic packaged-goods record (common type, no brand, no valid barcode, placeholder image) needs at least one piece of positive identifying evidence; the product-type name alone is insufficient. Brand, product type, and size must agree; sub-brands, special editions, and bundles are distinct products; if GTIN \emph{or} UPC matches, treat the barcode as matching. For meat, fish, and produce, origin, cut, and specific type matter, and generic names can still match.\\[2pt]
\textit{Evidence gathering.} Search the open web on the barcode, name, or any field to confirm brand, size, or variant when the records are inconclusive.\\[2pt]
\textit{Output.} JSON \{\texttt{reasoning}, \texttt{match\_decision}\}, with no confidence score.\\[2pt]
\textit{Open-weight addenda.} Count only same-retailer evidence as identity proof; treat empty or noisy search results as inconclusive rather than as proof a barcode is fabricated; cap the tool loop at four rounds.
\end{minipage}}
\caption{The shared base prompt (abridged) plus the open-weight model's addenda. The full rule set also covers null handling, size and unit conversion, image comparison, and bundles.}
\label{fig:prompt}
\end{figure}

\subsubsection{Migrating from a Closed to an Open-Weight Stack}
The production system replaces the closed frontier VLM (GPT-5.4) plus its native, OpenAI-hosted web-search tool with a self-hosted, open-weight Mixture-of-Experts VLM, Qwen 3.6 35B-A3B (35B total and 3B active parameters, FP8-quantized, served on H200-class GPUs). It is paired with an internally-maintained web-search tool behind the MCP interface. The reasoning model and the search backend are separated behind a stable tool contract, so the search backend can be swapped without touching the agent's prompt or output schema; this decoupled search-grounding design mirrors recent gateway-style architectures \cite{dsg2026}. Moving to open weights also removes per-token vendor pricing and lets us own the throughput and latency envelope. Replacing a closed API with a cheaper open model at comparable quality is an established pattern for language models~\cite{scalingdown2024slm}, now extending to agentic tool use~\cite{agenticqwen2026}; unlike those systems, the migration required no model training, and the open-weight agent is driven entirely by a shared base prompt plus a small set of model-specific addenda (Figure~\ref{fig:prompt}).

\subsubsection{Taming the Agentic Loop}
An unconstrained agent tends to over-search, spending tool calls that add latency and cost without new evidence. We cap the agent at four tool rounds, which is the empirical F1 and precision peak: rounds five through twenty re-issue searches that return already-seen information rather than resolving evidence, so the cap trims cost with no accuracy loss. Two prompt addenda close most of the open-weight gap. A grounding addendum restricts what counts as sufficient grounding for a barcode or brand claim to same-retailer evidence only. An interpretation addendum instructs the model to treat empty or noisy search results as inconclusive rather than as positive evidence that a barcode is fabricated. On weighted (non-barcoded) items, where searches frequently return little, this single change recovers 3~pp of recall. Together these addenda turn a general-purpose open model into a matcher that behaves like the closed baseline.

\subsection{Routing}
\label{sec:routing}
Adaptive routing composes the three stages, deciding from the cross-encoder's score whether each pair is resolved cheaply or escalated. Stage~2 emits a calibrated score $s \in [0,1]$ per pair and applies two thresholds that partition the score axis into three bands. Scores at or above the HIGH threshold are auto-accepted as links. Scores at or below the LOW threshold are auto-rejected. The residual MEDIUM band is escalated to Stage~3. Thresholds are placed to protect precision and recall independently. The HIGH threshold is the lowest score whose implied precision is at least 98\%. The LOW threshold is the highest score that still admits at least 98\% of positives above it. MEDIUM is whatever remains.

\paragraph{Expected cost.} Under this contract the expected per-pair cost is
\begin{equation}
\mathbb{E}[c] = c_{\text{cheap}} + p(\text{MEDIUM}) \cdot c_{\text{expensive}},
\label{eq:cost}
\end{equation}
where $c_{\text{cheap}}$ is paid on every pair and the expensive Stage-3 term is paid only on the escalated fraction $p(\text{MEDIUM})$. Because $c_{\text{expensive}} \gg c_{\text{cheap}}$ (Table~\ref{tab:ladder}), the escalation rate, not the raw per-call cost, dominates the aggregate spend, so narrowing the MEDIUM band is the primary cost lever.

\begin{table}[t]
\centering
\caption{Relative per-pair cost across the three matching models, normalized to the cheap cross-encoder ($1\times$); values are relative multipliers only.}
\label{tab:ladder}
\begin{tabular}{@{}lll@{}}
\toprule
Model & Rel.\ cost & Modality \\
\midrule
Cross-encoder matcher & $1\times$ & text \\
Open-weight agentic VLM & $\sim$$7{,}000\times$ & text{+}image{+}search \\
Closed frontier VLM & $\sim$$50{,}000\times$ & text{+}image{+}search \\
\bottomrule
\end{tabular}
\end{table}

The ladder spans nearly five orders of magnitude, from the text-only cross-encoder to the closed frontier VLM. Each open-weight agent call costs roughly $7{,}000\times$ a cross-encoder call, and the closed VLM is $7\times$ costlier still at equal precision. This gap is why the cascade routes the easy majority away from Stage~3.

\section{Evaluation}
\label{sec:evaluation}

We report Stage 1's retrieval coverage directly, then evaluate Stage 2's matcher accuracy and routing, and Stage 3 against both trained human operators and the closed model it replaces.

\subsection{Stage 1: Retrieval Recall}
\label{sec:stage1eval}

The evaluation set combines two families of ground truth that behave differently. The linked-pairs slice evaluates records against their currently linked canonical product, i.e., pairs the production retriever already surfaced when that link was first made; recall here is partly self-reinforcing, a non-regression check rather than a test of discovery power. The deduplication slice consists of pairs of canonical products independently identified as duplicates through a separate effort, and was largely not previously surfaced by the current retriever, so its recall carries no such circularity and is the more informative signal of genuine retrieval headroom. One caveat cuts the other way: crediting a since-merged product's canonical successor as correct (Section~\ref{sec:overview}) was necessary for 64\% of deduplication-slice pairs versus 5\% of linked-pairs-slice pairs, so the harder slice leans more heavily on that adjustment. We report both figures and treat deduplication-slice recall as the primary metric below, with the combined figure as secondary context. On production traffic, the retrieval configuration of Section~\ref{sec:retrieval} (image, text, and GTIN ANN search, merged by unweighted Reciprocal Rank Fusion, RRF, into a final $K{=}20$) reaches 93.06\% recall on the primary deduplication-slice metric versus 99.12\% on the linked-pairs slice (96.89\% combined). Retrieval recall remains a ceiling we track separately from matcher accuracy (Section~\ref{sec:overview}).

We evaluated each retrieval lever of Section~\ref{sec:retrieval} directly; Table~\ref{tab:retrieval} reports deduplication-slice recall for each single-factor change to the production configuration. Two levers were unambiguous wins at no added downstream cost: widening the per-channel candidate pool while holding the final $K{=}20$ fixed, and adding a lexical BM25 channel over item names alongside the existing embedding and barcode channels (BM25 alone reached 88.6\% deduplication-slice recall at $K{=}20$ and 94.61\% at $K{=}50$). Replacing the general-purpose text embedding with a stronger, jointly-trained multimodal embedding (Gemini Embedding 2) was the largest single win on the primary metric. A comparably-sized open alternative (Qwen3-VL) needed heavy product-specific prompt engineering before its embeddings were usable at all and, even after tuning, underperformed the baseline once fused, with most of the drop concentrated in the linked-pairs slice; an untuned embedding swap thus risks breaking already-solved pairs more than it adds discovery power.

Two further levers, evaluated as add-ons to the strongest single-factor configuration (Gemini Embedding 2), looked negative on the combined metric; only one remains negative on the primary metric. A dedicated UPC channel lowered recall on both metrics (94.27\%$\to$94.05\% primary, 97.42\%$\to$97.26\% combined): UPC and GTIN candidates are correlated and similarly noisy, so a second barcode-like channel displaced stronger candidates from the fixed final pool, despite contributing a few unique true positives on its own. A coarse-category prefilter, by contrast, was within noise on the primary metric (94.27\%$\to$94.12\%, and 93.06\%$\to$93.16\% when applied to the production baseline instead) even though it looked like a regression on the combined figure (97.42\%$\to$97.07\%). Nearly all of that drop traces back to the linked-pairs slice (99.24\%$\to$98.78\%), so we read the category prefilter as roughly neutral for true discovery but still worth guarding against as a regression risk on already-linked pairs. Coverage is not the binding constraint: a union-ceiling analysis puts the fraction of pairs whose ground truth appears in some channel's top-50 at 99.03\% combined, only two points above the current fused recall, so most achievable recall is already in the retrieved pool and the remaining gap is dominated by reranking rather than channel coverage.

Isolating each channel on the primary metric sharpens this picture. The image channel alone reaches only 0.40\% recall on the deduplication slice (1\% combined) and contributes zero unique ground-truth hits after fusion: every match it recovers is also found by text or barcode search. Its vision encoder reliably recognizes a product's visual family, packaging and brand, but cannot read the discriminative on-package text that separates near-identical variants, a limited-edition flavor from the standard one, for instance. The GTIN channel reaches 73\% recall on this harder slice (82\% combined), while text search, lexical or dense, remains the strongest single channel at 87--92\% depending on the encoder (low-to-mid 90s combined). Qualitative review of the remaining misses attributes a meaningful share to catalog noise rather than weak retrieval: duplicate canonical products competing for the same query and, in a minority of cases, a mislabeled ground truth, echoing the catalog-quality pressure discussed in Section~\ref{sec:discussion}.

\begin{table*}[t]
\centering
\caption{Retrieval recall on the deduplication slice (the primary, independent-ground-truth metric; Section~\ref{sec:stage1eval}), under the production configuration (image+text+GTIN ANN, RRF fusion, $K{=}20$) and single-factor changes to it.}
\label{tab:retrieval}
\begin{tabular}{@{}lc@{}}
\toprule
Configuration & Recall \\
\midrule
Baseline (production) & 93.06\% \\
+ deeper per-channel candidates, fixed final $K$ & 93.38\% \\
+ lexical (BM25) channel & 94.10\% \\
+ stronger multimodal embedding (Gemini Embedding 2) & 94.27\% \\
+ open multimodal embedding (Qwen3-VL) & 92.58\% \\
\bottomrule
\end{tabular}
\end{table*}

\subsection{Stage 2: Matcher Accuracy}
\label{sec:stage2eval}

\subsubsection{Model Discovery via Autoresearch}
\label{sec:stage2disc}
Rather than hand-tune the recipe, we ran model discovery as an ``autoresearch'' loop in which AI agents autonomously proposed, launched, and evaluated experiments with humans in the loop, 43 in total, following an emerging pattern of AI agents that plan, run, and analyze machine-learning experiments, studied in academic settings~\cite{dolphin2025autoresearch,aiscientistv2} and applied to production ML with a human in the loop~\cite{tripcom2026coscientist,kuaishou2026agentx}. Measured by R@P98 (recall at the $\geq 98\%$ precision bar Stage 2 uses for its HIGH-band auto-accept threshold, Section~\ref{sec:routing}), the clearest finding is that at this operating point label volume matters far more than model capacity: scaling the consensus-labeled training set to 5.3M pairs improved the operating point more than any architectural or tuning change we tried, while a larger backbone and a dedicated hyperparameter sweep did not help. A coarse category tag was the one clearly worthwhile input feature; the barcode representation mattered most of all, but carried an unexpected failure mode (Section~\ref{sec:stage2gtin}).

\subsubsection{Barcode Over-Indexing and Its Failure Mode}
\label{sec:stage2gtin}

Feeding raw barcode digit tokens directly into the sequence, rather than a precomputed \texttt{[BARCODE\_MATCH]} equality flag, was the single largest representation win we found on held-out data ($+0.8$pp F1). The model learns partial and noisy barcode agreement from the tokens themselves rather than collapsing that signal into one bit. This win has a failure mode the held-out set does not exercise. Some canonical products carry a barcode that has been reused or corrupted upstream, a dirty or duplicated identifier shared across otherwise-unrelated products. A matcher trained to trust raw barcode agreement over-indexes on such cases, producing a wrong link that scores high confidence whenever a merchant record's item name plainly conflicts with the canonical product's name but the barcodes happen to agree. Barcode-collision pairs of this kind are rare enough in the main held-out set that ordinary evaluation does not surface the risk; a stress audit built specifically to over-sample same-barcode, conflicting-name pairs does.

We address this with a second layer of distillation rather than by discarding the raw-digit representation, which remains the best signal on the great majority of pairs where barcodes are clean. A first mitigation regularizes training with barcode dropout and a name-similarity guardrail. Barcode dropout randomly withholds the barcode field during training so the model cannot over-rely on it; the guardrail caps the training target whenever two records share a barcode but their name embeddings disagree. Genuine barcode-collision pairs are too rare in the natural training distribution to teach this behavior on their own, so we synthesize additional collision pairs by deliberately pairing a real barcode with a mismatched product name. This gives the model enough exposure to the failure mode to learn to discount it. The best recipe found so far treats this regularized model as a teacher: its guardrail-corrected scores become soft training targets for a fresh student trained for two epochs on a rebalanced, collision-cleaned corpus. This folds the fix into a single deployable model with no extra runtime component. Table~\ref{tab:gtincollision} tracks recall at the 98\% precision operating point on the collision-stress audit across this progression. The fix is not yet re-validated on a fresh, non-adversarial audit, so we report it as the current best candidate rather than a closed result.

\begin{table*}[t]
\centering
\caption{Recall at the 98\% precision operating point (R@P98) on a stress audit oversampled for barcode-collision pairs (same barcode, conflicting name), across the mitigation progression. Single-seed for the distillation rows (Section~\ref{sec:stage2gtin}).}
\label{tab:gtincollision}
\begin{tabular}{@{}lc@{}}
\toprule
Recipe & R@P98 (stress audit) \\
\midrule
Raw barcode digits, no mitigation & 0.200 \\
+ barcode dropout + name-similarity guardrail & 0.692 \\
+ distillation from the guardrailed teacher, 1 epoch & 0.743 \\
+ distillation from the guardrailed teacher, 2 epochs & 0.768 \\
\bottomrule
\end{tabular}
\end{table*}

A representation choice that wins decisively in aggregate can still carry a concentrated, rare-but-severe failure mode that only a deliberately adversarial evaluation slice will find: no single evaluation slice, however large, is sufficient evidence of robustness on its own. The fix is also a second instance of the paper's central distillation idea (Section~\ref{sec:stage2}): rather than hand-labeling collision cases, a regularized teacher's own corrected judgments become the training signal for a cleaner student.

\subsubsection{Results}
\label{sec:stage2results}
On the zero-leak held-out set (6k pairs, 38\% positive) the matcher reaches AP 0.964, best F1 0.896, and R@P98 of 77.05\% (single-seed; run-to-run noise of $\pm 1.5$ pp). Because this held-out set is near-balanced (38\% positive), its R@P98 is not directly comparable to the HIGH-band auto-accept share reported on the low-prevalence production audit below.

An independent, ops-certified audit (24k pairs drawn from a production candidate-linking run, each pair scored first by the Stage-3 VLM and then validated or corrected by human operators) grounds these results in human judgment. Against the certified labels, the matcher auto-accepts 43.7\% of links in its HIGH band at the 98\% precision bar, and it routes 95\% of barcode-matched true links to HIGH. We also verified that BF16 GPU inference is numerically equivalent to FP32 CPU inference (band agreement 99.95\% over 32.9M rows), so the same matcher can serve the online build, offline post-build, and a separate inventory-ingestion pipeline despite their differing substrates. This decouples the model from any single serving substrate.

\subsection{Stage 3: Hard-Tail Agent Performance}
\label{sec:stage3eval}

\subsubsection{Accuracy Versus Trained Human Operators}
\begin{figure}[t]
\centering
\fbox{\begin{minipage}{0.90\columnwidth}
\footnotesize
\textbf{Agentic web search on a hard pair.}\\[3pt]
\textbf{Records.} Two ``branded six-quart Dutch oven'' listings, near-identical names and the same product type; an operator and an independent auditor both marked them a match.\\
\textbf{Missing evidence.} The deciding attribute is in neither record.\\
\textbf{Agent.} A web search on the barcode shows the record's product is \emph{tri-ply stainless steel}, whereas the candidate is \emph{enameled cast iron}.\\
\textbf{Verdict.} \texttt{match\_decision: No}, a different product; neither reviewer had checked the barcode.
\end{minipage}}
\caption{A representative hard-tail case: the text-only matcher escalates the pair, and the agent recovers the deciding evidence (product material), absent from both catalog records, through open-web search.}
\label{fig:example}
\end{figure}
The prior-generation agent (a closed frontier VLM, GPT-5.4) established that giving a VLM the same image-plus-search access as a human operator (Section~\ref{sec:stage3}) outperforms human review rather than just matching it. Ground truth for this comparison was set to avoid favoring either side: operators and the agent labeled the same pairs independently, agreements were taken as correct, and the pairs on which they disagreed were adjudicated by an independent reviewer panel drawn from a different group than the original labelers. Figure~\ref{fig:example} shows a representative case, where open-web search on a barcode overturned a match both a human operator and an auditor had accepted. On the hardest medium-confidence review tasks, the agent outperformed trained operators, improving accuracy by 13.7~pp, recall by 18.5~pp, and precision by 4.7~pp, all at $p<0.0001$ (Table~\ref{tab:human}), and reached 99.15\% precision at 95.47\% recall. Because these gains land on precisely the pairs the upstream matcher could not resolve, they translate into higher automatic coverage and fewer wrong links surviving into the live catalog. This result sets the accuracy bar for the stage: any cost-reduction effort must preserve it. The precision and recall reported elsewhere (Table~\ref{tab:migration}) come from a separate open-versus-closed audit on a different sample, so those absolute numbers are not directly comparable to these.

\begin{table*}[t]
\centering
\caption{Agentic VLM versus trained human operators on the hardest medium-confidence match tasks. Differences are significant at $p<0.0001$; the 95\% confidence intervals on the deltas exclude zero.}
\label{tab:human}
\begin{tabular}{@{}lccc@{}}
\toprule
System & Accuracy (\%) & Precision (\%) & Recall (\%) \\
\midrule
Human operators & 83.0 & 94.4 & 77.0 \\
Closed VLM (GPT-5.4) & 96.7 & 99.1 & 95.5 \\
\midrule
$\Delta$ (VLM $-$ operators) & $+13.7$ & $+4.7$ & $+18.5$ \\
\bottomrule
\end{tabular}
\end{table*}

\subsubsection{End-to-End Production Coverage}
On a production sample of roughly two million merchant records, the cheap stage alone linked 68.1\% of records at the operating precision bar (record-level coverage, distinct from the pair-level auto-accept share of Section~\ref{sec:stage2results}), and escalating the residual band to the agent raised end-to-end coverage to 77.1\%, a gain of 9.0 percentage points. The uplift is large precisely because it falls on the most ambiguous records, those the cheap stage cannot resolve, where each additional correct link is hardest to win. Coverage does not approach 100\%: recall is bounded at every stage, and the global catalog does not hold a canonical entry for every product on the market, so some records have no correct target to link to. This sample was served by the previous-generation cheap linker; the distilled cross-encoder of Section~\ref{sec:stage2} replaces that stage at a fraction of the per-pair cost, and the agent's uplift is a property of the escalated tail rather than of the particular cheap model that routes into it.

\subsubsection{Cost and Latency: Open Versus Closed}
At equal 98.0\% precision, the open-weight agent is 7$\times$ cheaper per pair than the closed stack, at a recall of 88\% versus 92\% (Table~\ref{tab:migration}). Two levers drive the reduction: self-hosting the open weights removes per-token vendor pricing, and an internally-maintained web-search tool is cheaper than OpenAI's hosted web search. Decoupling the search backend from the reasoning model is what lets us swap in the cheaper tool. The self-hosted deployment scales to zero when idle and sustains the escalation volume on a handful of GPUs, at the cost of a 1--3~min cold start. Because the upstream matcher already absorbs most traffic, this stage handles only the small escalated volume.

\begin{table*}[t]
\centering
\caption{Open-weight self-hosted agent versus the closed frontier agent on the hard-tail matching stage, at equal precision. Cost is per pair, normalized to the open-weight stack (open $=1\times$ here; Table~\ref{tab:ladder} gives the absolute cost ladder); other rows are relative or architectural.}
\label{tab:migration}
\begin{tabular}{@{}lll@{}}
\toprule
Dimension & Open (Qwen 3.6, self-hosted) & Closed (GPT-5.4, API) \\
\midrule
Precision & 98.0\% & 98.0\% \\
Recall & 88\% & 92\% \\
Relative cost/pair & $\sim$1$\times$ & $\sim$7$\times$ \\
Search backend & self-hosted web search (MCP) & OpenAI-hosted web search \\
Fine-tuning / weights & full control & none \\
\bottomrule
\end{tabular}
\end{table*}

\section{Discussion}
\label{sec:discussion}

\subsection{Labels as the Bottleneck}
The accuracy ceiling of the cascade was set by labels, not by architecture. The distilled matcher is trained on judgments the cascade generates about itself: the expensive Stage-3 multimodal reasoner adjudicates hard pairs, and those adjudications become supervision for the cheap Stage-2 cross-encoder \cite{hinton2015distilling}. The marketplace thus pays the frontier-reasoning cost once and amortizes it across every pair the cheap Stage-2 model later resolves (Section~\ref{sec:stage2}).

Beyond volume, label quality drove the result. A single automated labeler injects idiosyncratic noise that a student model will memorize. We instead used dual-VLM consensus (Section~\ref{sec:stage2}): keeping only pairs on which two independent VLMs agree trades a modest amount of coverage for higher label consistency, and turns the disagreement rate itself into a usable signal of pair difficulty. This raises a circularity risk: if VLMs grade VLM-derived labels, the whole evaluation could be internally consistent but wrong. The independent ops-certified audit (Section~\ref{sec:stage2results}) breaks that circularity. Operators corrected the VLM where it erred and otherwise agreed closely, which grounds the distilled labels in human judgment without paying for human labels at training scale.

Given clean labels, data volume dominated every architectural lever we tried (Section~\ref{sec:stage2disc}): scaling the label set moved the operating point far more than model size or a dedicated hyperparameter sweep. The practical implication is a reordering of effort: for production entity resolution \cite{christen2012,konda2016magellan,peeters2024wdc}, label-generation pipelines deserve more investment than model search. The autoresearch process (AI agents planning, launching, and evaluating experiments under human oversight) mattered because it made that search cheap enough to establish the data-beats-architecture result empirically rather than by assertion \cite{buildjudgeoptimize2026}.

\subsection{Open Source vs.\ Closed Source: Cost, Latency, Control}
Replacing the closed frontier VLM with a self-hosted open-weight Mixture-of-Experts VLM was an economic and operational decision as much as an accuracy one (Table~\ref{tab:migration}, Section~\ref{sec:stage3eval}).

Idle cost is zero, since the service scales to zero between bursty offline dedup workloads. Owning the web-search tool rather than metering OpenAI's hosted web search, together with caching and batching searches, is a direct cost lever \cite{dsg2026}. Ownership of the weights also unlocks levers a closed API denies us: prompt addenda tuned to our search evidence and the option to fine-tune later. However, the closed model still wins narrowly on recall (Section~\ref{sec:stage3eval}).

\subsection{A Hard Instance of a Generic Pipeline}
Retrieve-then-match entity resolution is textbook \cite{li2020ditto,christen2012}; this deployment is a hard instance of it. Scale is the first pressure (Section~\ref{sec:overview}): a retrieval stage that multiplies each already-massive record set into $K$ further candidate pairs. That multiplication is why the cascade must route by difficulty \cite{violajones2001,chow1970reject}, spending multimodal reasoning only on the ambiguous MEDIUM band.

Catalog quality is the second pressure, and it shaped the design directly. Merchant-supplied data carries missing, placeholder, or wrong images, low and un-normalized barcode coverage, size in free text, and invalid structured fields. Each design choice is a response: feeding raw barcode digits rather than a match flag lets the model exploit partial and noisy identifiers; the multimodal Stage~3 agent stays usable when one side's image is a placeholder or missing, reasoning from whatever image and text are present; and agentic web search fills in missing information at the hard tail. Product names span many languages and scripts, which we treat qualitatively and flag as an evaluation gap. Multi-category heterogeneity (many top-level categories, fresh vs.\ packaged, weighted vs.\ each, barcodes reliable for packaged goods but not meat or produce) motivates the coarse category tag \cite{warner2024modernbert}. Finally, the same matcher serves the online build, offline post-build dedup, and a separate inventory-ingestion pipeline with a different schema and latency regime. This is why the GPU-equals-CPU numerical equivalence of Section~\ref{sec:stage2eval} matters: it lets us place the identical model wherever each pipeline's constraints dictate.

\section{Conclusion}

We described a production entity-resolution system for a large multi-category on-demand marketplace that frames catalog linking as retrieve-then-match and spends compute in proportion to difficulty. A confidence-routed cascade auto-resolves the high-signal majority with a distilled text-only cross-encoder and escalates only the ambiguous middle to an agentic multimodal VLM that can search for missing information. This raises end-to-end link coverage while reducing operator workload and wrong links. Distillation serves two purposes: the cheap stage inherits the expensive stage's judgment, and, when a representation it relies on carried a hidden failure mode, a more careful teacher's corrected judgments retrained a cleaner student rather than hand-patching the model. The design is portable and cost-bounded: the escalation policy, the VLM-as-labeler distillation loop, and the open-weight self-hosted reasoning tier are domain-independent, and each stage can be swapped as models improve; the domain itself (noisy merchant catalogs, unreliable barcodes and images, multilingual names, heterogeneous categories) does not. Confidence-routed escalation lets a team spend expensive reasoning only where that difficulty concentrates.

\begin{acks}
We thank Carolyn Tang, Kavya Adimulam, and Jacob Eisenach for their operations
partnership: the curated ground truth and the operator-certified audits they
organized are what made the evaluations in this paper measurable rather than
anecdotal. We thank Mat Amorim for the analytics support behind the coverage and
quality sizing. We are grateful to Sudeep Das and Rohit Pooserla, whose ideation helped shape
the cascade design and whose sponsorship made this work possible, and to
Pingyang He for his support of the engineering work behind it.
\end{acks}

\section*{Declaration on Generative AI}

During the preparation of this work, the author(s) used generative AI tools in order to: Drafting content; Content enhancement; Paraphrase and reword; and Grammar and spelling check. After using these tool(s)/service(s), the author(s) reviewed and edited the content as needed and take(s) full responsibility for the publication's content.

\bibliographystyle{ACM-Reference-Format}
\bibliography{references_arxiv}

\end{document}